\documentclass[lettersize,journal]{IEEEtran}
\usepackage{amsmath,amsfonts}
\usepackage{algorithmic}
\usepackage{algorithm}
\usepackage{array}
\usepackage{subcaption}
\usepackage{textcomp}
\usepackage{stfloats}
\usepackage{url}
\usepackage{verbatim}
\usepackage{graphicx}
\usepackage{threeparttable}
\usepackage{cite}
\usepackage{multirow}
\usepackage{booktabs}
\usepackage{xcolor}
\usepackage[utf8]{inputenc}
\usepackage{bbm}
\usepackage{cite}

\begin{document}

\title{Attention-Enhanced Prototypical Learning for Few-Shot Infrastructure Defect Segmentation}

\author{Christina Thrainer,~\IEEEmembership{}
        Md Meftahul Ferdaus,~\IEEEmembership{}
        Mahdi Abdelguerfi,~\IEEEmembership{}
        Christian Guetl,~\IEEEmembership{}
        Steven Sloan,~\IEEEmembership{}
        Kendall N. Niles,~\IEEEmembership{}and~Ken Pathak~\IEEEmembership{}
        \thanks{M. Ferdaus and M. Abdelguerfi are with the Canizaro Livingston Gulf States Center for Environmental Informatics, the University of New Orleans, New Orleans, USA (e-mail: mferdaus@uno.edu; gulfsceidirector@uno.edu).}
        \thanks{C. Thrainer and C. Guetl are with the Graz University of Technology, Graz, Austria.}
        \thanks{S. Sloan, K. N. Niles, and K. Pathak are with the US Army Corps of Engineers, Engineer Research and Development Center, Vicksburg, Mississippi, USA.}
        \thanks{Manuscript received July XX, 2025; revised July XX, 2025.}}

\markboth{IEEE Transactions on Emerging Topics in Computing,~Vol.~XX, No.~X, July~2025}%
{Alshawi \MakeLowercase{\textit{et al.}}: Enhanced Feature Pyramid Networks for Few-Shot Semantic Segmentation in Infrastructure Defect Detection}

\maketitle

\begin{abstract}
Few-shot semantic segmentation is vital for deep learning-based infrastructure inspection applications, where labeled training examples are scarce and expensive. Although existing deep learning frameworks perform well, the need for extensive labeled datasets and the inability to learn new defect categories with little data are problematic. We present our Enhanced Feature Pyramid Network (E-FPN) framework for few-shot semantic segmentation of culvert and sewer defect categories using a prototypical learning framework. Our approach has three main contributions: (1) adaptive E-FPN encoder using InceptionSepConv blocks and depth-wise separable convolutions for efficient multi-scale feature extraction; (2) prototypical learning with masked average pooling for powerful prototype generation from small support examples; and (3) attention-based feature representation through global self-attention, local self-attention and cross-attention. Comprehensive experimentation on challenging infrastructure inspection datasets illustrates that the method achieves excellent few-shot performance, with the best configuration being 8-way 5-shot training configuration at 82.55\% F1-score and 72.26\% mIoU in 2-way classification testing. The self-attention method had the most significant performance improvements, providing 2.57\% F1-score and 2.9\% mIoU gain over baselines. Our framework addresses the critical need to rapidly respond to new defect types in infrastructure inspection systems with limited new training data that lead to more efficient and economical maintenance plans for critical infrastructure systems.
\end{abstract}


\section{Introduction}
\IEEEPARstart{A}{utomated} detection and segmentation of structural defects in civil infrastructure pose a significant challenge in computer vision, especially for underground water management systems like culverts and sewer pipes. When inspection is manual, it becomes laborious, time-consuming, and unreliable~\cite{kumar2020automated}. Traditional methods involve experts reviewing video from specialized inspection cameras, requiring significant human resources, inconsistent evaluation metrics, and subjective interpretation of defect severity~\cite{cha2017deep}. The emergence of automated semantic segmentation systems for detecting defects in infrastructure presents an opportunity for more accurate, consistent, and efficient inspection processes, improving ongoing maintenance strategies.

Applying deep learning to infrastructure inspection presents fundamental problems that separate it from a general computer vision task. First, obtaining labeled training data for infrastructure defects is costly and time-consuming, requiring specialized equipment, domain expertise, and unsafe working conditions in underground spaces~\cite{wang2022deep}. Second, the distribution of defect types in real-world infrastructure systems exhibits severe class imbalance, with common issues such as minor surface cracks being significantly overrepresented while critical but rare defects like structural collapses or major holes remain underrepresented in available training datasets~\cite{alshawi2024imbalance}. Third, the visual characteristics of infrastructure defects demonstrate substantial variability based on environmental conditions, pipe materials, structural age, and geographical factors, making it exceptionally difficult to develop models that generalize effectively across diverse operational contexts~\cite{guo2022attention}.

The Enhanced Feature Pyramid Network (E-FPN), originally developed for addressing class imbalance challenges in culvert and sewer defect segmentation, has demonstrated significant performance improvements over traditional semantic segmentation approaches through the incorporation of architectural innovations including sparsely connected blocks and depth-wise separable convolutions~\cite{alshawi2024imbalance}. The E-FPN architecture is well-suited to handle the multi-scale issue of infrastructure defects, since it uses a hierarchical pyramid of feature maps that can represent coarse global structural features, as well as detail-oriented defects at multiple resolutions. Nevertheless, while the E-FPN was effective for handling multi-scale representation and class imbalance associated with infrastructure inspection tasks, it still requires considerable labeled training data for best segmentation performance.

Few-shot learning has been a particularly beneficial paradigm for addressing data scarcity challenges \cite{ferdaus2025few} in targeted domains like infrastructure inspection. Rather than needing large amounts of labeled data for every class of their data as traditional supervised learning models do, few-shot learning approaches try to allow a model to classify and segment new classes with only a few different examples labeled~\cite{wang2020generalizing}. This is particularly relevant in an infrastructure inspection context, where new defects might exist because of changing environments, aging infrastructure systems, or newly implemented materials and construction strategies that proved to be absent in the previous datasets.

Prototypical networks represent one of the most successful and theoretically grounded approaches to few-shot learning, operating on the fundamental principle that each semantic class can be effectively represented by a prototype computed as the centroid of its support examples within a learned embedding space~\cite{snell2017prototypical}. The classification of query examples is subsequently performed based on distance computations to these class prototypes, typically employing cosine similarity or Euclidean distance metrics. This approach has demonstrated remarkable success in few-shot classification tasks and has been successfully extended to more complex problems such as semantic segmentation through sophisticated architectures like Prototype Alignment Network (PANet)~\cite{wang2019panet}.

The integration of prototypical learning mechanisms with robust multi-scale feature extraction architectures presents a significant opportunity to address both the data scarcity constraints and the performance requirements of practical infrastructure inspection systems. By combining the proven multi-scale feature extraction capabilities of E-FPN with the few-shot learning capabilities of prototypical networks, it becomes feasible to develop systems that can rapidly adapt to novel defect types with minimal additional training data while maintaining high segmentation accuracy across diverse operational scenarios.

Attention mechanisms have proven to be crucial components in modern deep learning architectures, enabling models to selectively focus on the most relevant features for specific tasks while suppressing irrelevant background information~\cite{vaswani2017attention}. In the context of few-shot semantic segmentation, attention mechanisms can significantly enhance the quality of prototype generation by emphasizing discriminative features and reducing the influence of noisy or irrelevant background patterns. Different types of attention mechanisms, including self-attention, local self-attention, and cross-attention, offer varying approaches to feature enhancement and prove particularly valuable in scenarios where the distinction between different defect types relies on subtle visual differences that may be easily overlooked by standard convolutional operations.

This paper presents a comprehensive investigation into the adaptation of E-FPN for few-shot semantic segmentation of infrastructure defects through prototypical learning mechanisms. Our approach systematically addresses several critical research questions.
\begin{itemize}
\item How can the robust multi-scale feature extraction capabilities of E-FPN be effectively integrated with prototypical learning frameworks for few-shot scenarios?
\item What constitutes the optimal training strategy for balancing the benefits of pre-trained feature representations with the specific requirements of few-shot adaptation?
\item How do different attention mechanisms contribute to prototype quality and overall segmentation performance in few-shot learning settings?
\item What are the practical implications and deployment considerations for real-world infrastructure inspection applications?
\end{itemize}

Our primary contributions can be summarized as follows: First, we develop a novel architecture that seamlessly integrates the E-FPN encoder with a sophisticated prototypical learning framework specifically designed for few-shot semantic segmentation of infrastructure defects. Second, we propose and validate a two-stage training strategy that effectively leverages pre-trained E-FPN features while enabling efficient adaptation to few-shot scenarios through prototypical learning mechanisms. Third, we conduct a comprehensive evaluation of three distinct attention mechanisms and systematically analyze their impact on few-shot segmentation performance across various experimental configurations. Fourth, we provide extensive experimental validation demonstrating the effectiveness of our approach across different few-shot learning scenarios, with particular emphasis on the practical requirements and constraints of real-world infrastructure inspection applications.

The remainder of this paper is organized as follows: Section II provides a comprehensive review of related work in few-shot learning methodologies, semantic segmentation architectures, and infrastructure inspection applications. Section III details our proposed methodology, including the architectural design principles, training strategies, and attention mechanisms. Section IV presents our experimental setup, evaluation protocols, and performance metrics. Section V discusses the comprehensive results and their implications for practical infrastructure inspection applications. Section VI explores future research directions and potential extensions of our work, concluding with final remarks in Section VII.

\section{Related Work}

\subsection{Infrastructure Inspection and Automated Defect Detection Systems}

The application of computer vision techniques to infrastructure inspection has gained substantial attention due to the critical importance of maintaining aging infrastructure systems and the inherent limitations of traditional manual inspection methodologies~\cite{arafin2024deep}. Infrastructure defects present unique and challenging problems for automated detection systems due to their diverse visual characteristics, varying scales ranging from microscopic cracks to large structural failures, and the complex environmental conditions in which they occur, including poor lighting, debris obstructions, and inconsistent imaging perspectives.

Traditional approaches to infrastructure defect detection relied heavily on hand-crafted features and classical machine learning techniques, including edge detection algorithms, texture analysis methods, and support vector machines~\cite{zhou2022vision}. These methods often struggled with the substantial variability in defect appearance and the complex backgrounds typical of infrastructure environments, leading to high false positive rates and missed detections of critical defects. The advent of deep learning has fundamentally revolutionized this field, enabling more robust and accurate defect detection systems capable of handling the complexity and variability inherent in real-world infrastructure inspection scenarios.

Convolutional Neural Networks (CNNs) have been widely adopted for infrastructure defect detection, with various architectural approaches being systematically explored for different types of infrastructure systems and defect characteristics~\cite{wang2024data}. U-Net and its numerous variants have proven particularly popular for infrastructure inspection applications due to their ability to provide precise pixel-level segmentation while maintaining computational efficiency suitable for practical deployment~\cite{ronneberger2015u,alshawi2023dual}. However, these approaches often require large amounts of labeled training data and may struggle with the severe class imbalance typical of infrastructure defect datasets, where background pixels significantly outnumber defect pixels.

The E-FPN was specifically developed to address the multifaceted challenges of infrastructure defect detection in severely imbalanced datasets~\cite{alshawi2024imbalance}. The E-FPN incorporates several architectural innovations specifically designed to improve feature extraction capabilities and handle object variations common in infrastructure inspection scenarios. Sparsely connected blocks enable efficient information flow while reducing computational complexity, making the architecture suitable for resource-constrained deployment scenarios. Depth-wise separable convolutions maintain representational power while significantly reducing the number of parameters, enabling deployment on mobile and edge computing platforms typical of field inspection applications.

Class imbalance represents one of the most significant and persistent challenges in infrastructure defect detection applications. Real-world infrastructure systems typically exhibit highly skewed distributions of defect types, with common issues such as minor surface cracks being dramatically overrepresented while critical but rare defects such as structural holes or complete pipe collapses remain severely underrepresented in available training datasets~\cite{alshawi2024imbalance}. This imbalance can lead to models that perform exceptionally well on frequent defects but fail catastrophically to detect rare but potentially catastrophic issues that pose the greatest risk to infrastructure integrity and public safety.

\subsection{Few-Shot Learning Fundamentals and Methodological Approaches}

Few-shot learning has evolved from early investigations in meta-learning and transfer learning to become a fundamental paradigm for addressing data scarcity challenges in machine learning applications across diverse domains~\cite{hospedales2021meta}. The core principle underlying few-shot learning methodologies is the ability to leverage prior knowledge and accumulated experience to rapidly adapt to novel tasks with minimal training data, a capability that proves particularly crucial in specialized domains where data collection is expensive, time-consuming, or potentially hazardous.

The development of few-shot learning can be systematically traced through several key methodological approaches, each offering distinct advantages and limitations. Optimization-based methods, exemplified by Model-Agnostic Meta-Learning (MAML)~\cite{finn2017model}, focus on learning optimal initialization parameters that can be rapidly fine-tuned for novel tasks through gradient-based optimization. These approaches have demonstrated considerable success across various domains but often require careful hyperparameter tuning and can be computationally expensive during both training and adaptation phases. Metric-learning approaches, including Siamese networks~\cite{koch2015siamese} and matching networks~\cite{vinyals2016matching}, learn embedding spaces where similarity comparisons can effectively distinguish between different classes based on learned distance metrics. These methods have shown particular promise in scenarios where the number of classes may vary significantly between training and testing phases.

Prototypical networks, introduced by Snell et al.~\cite{snell2017prototypical}, represent a particularly elegant and theoretically grounded solution to few-shot learning challenges. The fundamental insight behind prototypical networks is that each semantic class can be effectively represented by a prototype computed as the centroid of its support examples in a learned embedding space. Classification is subsequently performed by computing distances between query examples and class prototypes, typically using Euclidean distance or cosine similarity metrics. This approach offers several significant advantages: computational efficiency during inference, theoretical grounding in optimal Bayes classifiers under certain assumptions, and natural handling of variable numbers of support examples per class without architectural modifications.

\subsection{Few-Shot Semantic Segmentation Architectures and Methodologies}

Few-shot semantic segmentation represents a natural but significantly more complex evolution of few-shot learning principles applied to dense prediction tasks. The challenge of few-shot semantic segmentation is substantially more complex than few-shot classification due to the requirement for pixel-level accuracy and the need to handle complex spatial relationships between different semantic regions within images~\cite{zhang2019canet}. Early approaches to few-shot semantic segmentation often relied on straightforward adaptations of classification-based few-shot learning methods, but these approaches frequently failed to capture the spatial complexity and multi-scale nature inherent in segmentation tasks.

PANet~\cite{wang2019panet} represents a significant methodological advancement in few-shot semantic segmentation. PANet introduces a sophisticated bidirectional alignment strategy that not only compares query features to support prototypes but also aligns support features to query predictions through iterative refinement processes. This bidirectional approach helps to significantly reduce the impact of intra-class variation and improves the robustness of prototype-based matching in challenging visual scenarios. The architecture employs masked average pooling techniques to generate class-specific prototypes from support images, ensuring that prototypes capture the essential characteristics of each semantic class while minimizing the influence of background regions and irrelevant visual patterns.

Recent developments in few-shot semantic segmentation have increasingly focused on improving prototype quality and addressing the fundamental challenges of multi-scale object recognition in complex visual scenes. Attention mechanisms have emerged as crucial components in these efforts, enabling models to focus selectively on the most discriminative features for prototype generation and query matching processes~\cite{yang2020prototype}. Self-attention mechanisms, in particular, have shown considerable promise in capturing long-range dependencies and improving the quality of feature representations used in prototype computation, leading to more robust and discriminative prototypes.

\subsection{Attention Mechanisms in Computer Vision Applications}

Attention mechanisms have become fundamental components in modern computer vision architectures, enabling models to selectively focus on the most relevant features for specific tasks while suppressing irrelevant background information~\cite{guo2022attention}. The concept of attention in computer vision draws direct inspiration from human visual attention systems, which naturally focus on salient regions while suppressing irrelevant background information through sophisticated neural mechanisms. This selective processing capability has proven particularly valuable in complex visual tasks where relevant information may be distributed across different spatial locations or feature channels, requiring sophisticated integration mechanisms.

Self-attention mechanisms, popularized by the revolutionary Transformer architecture~\cite{vaswani2017attention}, enable models to capture long-range dependencies and global context information that may be crucial for accurate visual understanding and interpretation. In the context of semantic segmentation applications, self-attention mechanisms can help models understand complex relationships between different spatial regions and improve the consistency of segmentation predictions across entire images. The computational complexity of standard self-attention has led to various optimization strategies, including local self-attention mechanisms that restrict attention computation to local neighborhoods while maintaining the essential benefits of attention-based feature enhancement.

Cross-attention mechanisms enable sophisticated comparison and alignment of features from different sources, making them particularly valuable in few-shot learning scenarios where support and query features must be effectively compared and aligned~\cite{chen2021semantic}. In few-shot semantic segmentation applications, cross-attention can facilitate better alignment between support prototypes and query features, leading to more accurate segmentation predictions and improved generalization to novel classes. The design of effective cross-attention mechanisms requires careful consideration of feature dimensions, computational efficiency, and the specific requirements of the matching task.

Expanding on these advances, this study introduces an innovative integration of E-FPN and prototypical learning, specifically customized to detect infrastructure defects. Our approach addresses challenges like class imbalance, varying defect scales, and limited labeled data by combining E-FPN's multi-scale feature extraction with attention-enhanced prototypical methods. Using a two-stage training strategy, our framework achieves rapid adaptability to new types of defects while ensuring high segmentation accuracy, effectively meeting the practical demands of real-world infrastructure inspection scenarios.

\section{Problem Formulation and Methodological Framework}

Let $\mathcal{D} = \{(\mathcal{S}_i, \mathcal{Q}_i)\}_{i=1}^N$ represent a collection of $N$ few-shot learning episodes, where each episode consists of a support set $\mathcal{S}_i$ and a query set $\mathcal{Q}_i$. For an $n$-way $k$-shot learning scenario, the support set is defined as $\mathcal{S}_i = \{(x_{i,j}^s, y_{i,j}^s)\}_{j=1}^{nk}$, containing $k$ labeled examples for each of $n$ semantic classes. The corresponding query set is $\mathcal{Q}_i = \{(x_{i,j}^q, y_{i,j}^q)\}_{j=1}^{n_q}$, where $n_q$ represents the number of query examples to be segmented. Each image $x \in \mathbb{R}^{H \times W \times C}$ has spatial dimensions $H \times W$ and $C$ channels, while the corresponding segmentation mask $y \in \{0, 1, \ldots, K-1\}^{H \times W}$ assigns each pixel to one of $K$ semantic classes including background.

The objective is to learn a mapping function $f_\theta: \mathcal{X} \rightarrow \mathcal{Y}$ parameterized by $\theta$ that can effectively segment infrastructure defects in query images based on limited support examples. Given an E-FPN backbone $\phi_{EFPN}$ that extracts multi-scale features $\{F_l\}_{l=1}^L$ at $L$ different pyramid levels, where $F_l \in \mathbb{R}^{H_l \times W_l \times C_l}$, our goal is to integrate prototypical learning mechanisms to enable effective few-shot adaptation.

For each pyramid level $l$, we define the feature extraction process as:
\begin{equation}
F_l = \phi_{EFPN}^{(l)}(x; \Theta_l),
\end{equation}
where $\Theta_l$ represents the learnable parameters of the E-FPN at level $l$. The multi-scale features are subsequently processed through attention-enhanced prototypical learning mechanisms to generate class-specific prototypes and perform query segmentation.

The complete few-shot segmentation framework can be formulated as:
\begin{equation}
\hat{y}^q = f_\theta(x^q | \mathcal{S}) = \sigma(\mathcal{H}(\{\phi_{EFPN}^{(l)}(x^q)\}_{l=1}^L, \mathcal{P}(\mathcal{S}))),
\end{equation}
where $\mathcal{P}(\mathcal{S})$ represents the prototype generation function applied to the support set, $\mathcal{H}$ denotes the prototype matching and segmentation head, and $\sigma$ is the softmax activation function that normalizes outputs into class probability distributions.

\begin{figure*}[t]
  \centering
  \includegraphics[width=0.9\textwidth]{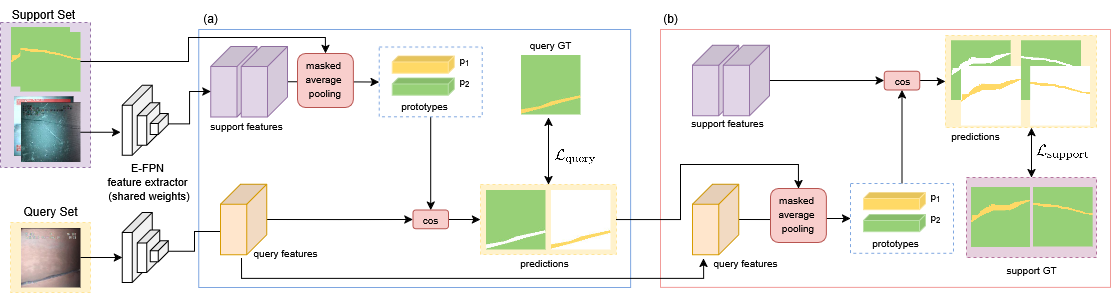}
\caption{Overall architecture of the proposed few-shot semantic segmentation framework. (a) Query-centric learning: prototypes generated from support features segment the query image, computing $\mathcal{L}_{query}$. (b) Support-centric learning: prototypes from query features segment support images, computing $\mathcal{L}_{support}$. The E-FPN encoder uses shared weights for both sets.}
  \label{fig:architecture}
\end{figure*}

\section{Proposed Enhanced Feature Pyramid Network for Few-Shot Semantic Segmentation}

We introduce an integrated model uniting the E-FPN with targeted prototypical learning techniques for effectively conducting few-shot semantic segmentation in identifying infrastructure defects. The framework is composed of three key elements: (1) a versatile E-FPN backbone for extracting multi-level features across scales, (2) prototypical learning modules enhanced by attention mechanisms to create and match robust prototypes, and (3) an advanced training procedure optimizing both pre-training and few-shot learning demands. 

As shown in Figure~\ref{fig:architecture}, our model merges E-FPN with a robust two-phase prototypical learning strategy to facilitate few-shot segmentation. This configuration is adept at deriving stable class prototypes from a small set of exemplar instances and extending this learning to new image queries. Initially, the E-FPN function extracts shared weight feature maps from both support and query datasets, ensuring quality, multi-scale representations. The core of our approach is a two-phase training routine. In phase one, presented in Figure~\ref{fig:architecture}(a), query-centric prototypical learning occurs. Prototypes are formed through masked average pooling on support features, yielding a powerful class representation for each semantic type. These are compared to query feature sets via cosine similarity to generate a detailed segmentation map for the query image. The query loss, $\mathcal{L}_{query}$, is determined by contrasting this forecast against the query's true labels. In phase two, outlined in Figure~\ref{fig:architecture}(b), we apply a support-centric approach to further polish the feature embeddings. Here, the query features shape prototypes that segment support images. These outputs are assessed against support ground truth, computing the support loss, $\mathcal{L}_{support}$. This dual-loss approach drives the feature extractor towards a more distinct and adaptive embedding capability, boosting the model's few-shot effectiveness.

\subsection{Enhanced Feature Pyramid Network Architecture}\label{fss_encoder}

Our E-FPN backbone incorporates several architectural innovations specifically designed for efficient multi-scale feature extraction in infrastructure inspection scenarios. The architecture consists of a bottom-up pathway with InceptionSepConv blocks and a top-down pathway with lateral connections for feature fusion across multiple scales.

\subsubsection{InceptionSepConv Block Design}
The InceptionSepConv block represents a key innovation in our architecture, combining the multi-scale processing capabilities of Inception modules with the parameter efficiency of depth-wise separable convolutions. Let $\mathbf{x} \in \mathbb{R}^{H \times W \times C}$ denote the input feature tensor with spatial dimensions $H \times W$ and $C$ channels. Each InceptionSepConv block processes input features through three parallel computational branches, formally defined as:

\begin{align}
\mathcal{B}_1(\mathbf{x}) &= \sigma(\gamma(\mathcal{D}_{3 \times 3}(\sigma(\gamma(\mathcal{C}_{3 \times 3}(\mathbf{x})))))) \label{eq:branch1} \\
\mathcal{B}_2(\mathbf{x}) &= \sigma(\gamma(\mathcal{D}_{5 \times 5}(\sigma(\gamma(\mathcal{C}_{5 \times 5}(\mathbf{x})))))) \label{eq:branch2} \\
\mathcal{B}_3(\mathbf{x}) &= \sigma(\gamma(\mathcal{C}_{1 \times 1}(\mathcal{M}_{3 \times 3}(\mathbf{x})))) ,\label{eq:branch3}
\end{align}
where $\sigma(\cdot)$ denotes the ReLU activation function, $\gamma(\cdot)$ represents batch normalization, $\mathcal{C}_{k \times k}(\cdot)$ denotes standard convolution with kernel size $k \times k$, $\mathcal{D}_{k \times k}(\cdot)$ represents depth-wise separable convolution with kernel size $k \times k$, and $\mathcal{M}_{k \times k}(\cdot)$ denotes max pooling with kernel size $k \times k$.

The complete InceptionSepConv transformation is formulated as the channel-wise concatenation of all branch outputs:
\begin{equation}
\mathcal{F}_{\text{ISC}}(\mathbf{x}) = \mathcal{B}_1(\mathbf{x}) \oplus \mathcal{B}_2(\mathbf{x}) \oplus \mathcal{B}_3(\mathbf{x}) ,\label{eq:isc_output}
\end{equation}
where $\oplus$ denotes the concatenation operation along the channel dimension.

This architectural design enables efficient capture of multi-scale feature representations while maintaining computational efficiency through the factorization of standard convolutions into depth-wise and point-wise operations, thereby reducing the total number of parameters and computational complexity.

\subsubsection{Multi-Scale Feature Pyramid Construction}
The feature pyramid construction follows a systematic hierarchical approach to capture semantic information at multiple resolution levels. Let $\mathbf{x}_0 \in \mathbb{R}^{H_0 \times W_0 \times C_0}$ denote the input image tensor. The bottom-up pathway generates a sequence of feature maps $\{\mathbf{C}_i\}_{i=1}^{5}$ at progressively reduced spatial resolutions:

\begin{align}
\mathbf{C}_1 &= \mathcal{F}_{\text{ISC}}^{(1)}(\mathbf{x}_0) \label{eq:c1} \\
\mathbf{C}_i &= \mathcal{F}_{\text{ISC}}^{(i)}(\mathcal{P}(\mathbf{C}_{i-1})), \quad i \in \{2, 3, 4, 5\} \label{eq:ci},
\end{align}
where $\mathcal{F}_{\text{ISC}}^{(i)}(\cdot)$ represents the $i$-th InceptionSepConv block and $\mathcal{P}(\cdot)$ denotes the pooling operation that reduces spatial dimensions by a factor of 2.

The top-down pathway constructs the enhanced feature pyramid $\{\mathbf{P}_i\}_{i=2}^{5}$ through lateral connections and feature fusion operations:

\begin{align}
\mathbf{P}_5 &= \mathcal{C}_{1 \times 1}^{(5)}(\mathbf{C}_5) \label{eq:p5} \\
\mathbf{P}_i &= \mathcal{C}_{1 \times 1}^{(i)}(\mathbf{C}_i) + \mathcal{U}(\mathbf{P}_{i+1}), \quad i \in \{4, 3, 2\} \label{eq:pi},
\end{align}
where $\mathcal{C}_{1 \times 1}^{(i)}(\cdot)$ denotes the lateral $1 \times 1$ convolution at level $i$ for channel dimension alignment, $\mathcal{U}(\cdot)$ represents the upsampling operation that doubles spatial dimensions through bilinear interpolation, and $+$ denotes element-wise addition.

This hierarchical feature extraction methodology enables the network to simultaneously capture fine-grained defect details at higher spatial resolutions and broader contextual semantic information at lower resolutions, facilitating robust multi-scale defect detection and segmentation across diverse infrastructure inspection scenarios.

\subsection{Prototypical Learning Framework for Few-Shot Segmentation}

Our prototypical learning framework adapts the principles of prototypical networks for dense prediction tasks in few-shot semantic segmentation. The framework consists of prototype generation, attention-enhanced feature processing, and prototype matching components.

\subsubsection{Masked Average Pooling for Prototype Generation}
Given support images and their corresponding segmentation masks, we generate class-specific prototypes through masked average pooling operations. For each semantic class $c$, the prototype is computed as:

\begin{equation}
p_c = \frac{1}{|\mathcal{S}_c|} \sum_{(x,y) \in \mathcal{S}_c} \frac{\sum_{h,w} f(x)_{h,w} \cdot \mathbf{1}[y_{h,w} = c] + \epsilon}{\sum_{h,w} \mathbf{1}[y_{h,w} = c] + \epsilon},
\end{equation}
where $f(x)$ represents the feature map extracted by the E-FPN encoder, $\mathcal{S}_c$ denotes the subset of support examples containing class $c$, $\mathbf{1}[\cdot]$ is the indicator function, and $\epsilon$ is a small constant to prevent division by zero.

This masked average pooling ensures that prototypes capture the essential characteristics of each defect class while minimizing the influence of background regions and irrelevant visual patterns.

\subsubsection{Attention-Enhanced Feature Processing}
We investigate three distinct attention mechanisms to enhance feature representations for improved prototype generation and matching:

\textbf{Self-Attention Mechanism:} The self-attention mechanism enables the model to capture long-range dependencies within feature maps:
\begin{align}
\text{SelfAttn}(F) &= \text{softmax}\left(\frac{QK^T}{\sqrt{d_k}}\right)V \\
Q &= FW_Q, \quad K = FW_K, \quad V = FW_V,
\end{align}
where $W_Q$, $W_K$, and $W_V$ are learnable projection matrices.

\textbf{Local Self-Attention Mechanism:} Local self-attention restricts attention computation to local neighborhoods, reducing computational complexity while maintaining spatial coherence:
\begin{equation}
\text{LocalAttn}(F)_{i,j} = \sum_{(p,q) \in \mathcal{N}(i,j)} \alpha_{(i,j),(p,q)} \cdot F_{p,q},
\end{equation}
where $\mathcal{N}(i,j)$ represents the local neighborhood around spatial position $(i,j)$.

\textbf{Cross-Attention Mechanism:} Cross-attention enables sophisticated alignment between support and query features:
\begin{equation}
\text{CrossAttn}(F^q, F^s) = \text{softmax}\left(\frac{F^q (F^s)^T}{\sqrt{d}}\right) F^s.
\end{equation}

\subsubsection{Prototype Matching and Segmentation}
The segmentation prediction for query images is computed using cosine similarity between query features and class prototypes:

\begin{equation}
P(y_{h,w}^q = c | x^q) = \frac{\exp(\alpha \cdot \cos(f(x^q)_{h,w}, p_c))}{\sum_{c'} \exp(\alpha \cdot \cos(f(x^q)_{h,w}, p_{c'}))},
\end{equation}
where $\alpha$ is a learnable temperature parameter that controls the sharpness of the probability distribution, empirically set to 20 in our experiments.

\subsection{Training Strategy and Optimization Framework}

Our training strategy employs a two-stage approach that balances the benefits of pre-trained feature representations with the specific requirements of few-shot adaptation.

\subsubsection{Stage 1: Encoder Pre-training}
In the first stage, we pre-train the E-FPN encoder using the complete dataset with a non-trainable prototypical head. This stage focuses on learning robust multi-scale feature representations that capture the essential characteristics of infrastructure defects across different scales and environmental conditions.

The pre-training objective combines multiple loss components:
\begin{equation}
\mathcal{L}_{\text{pretrain}} = \alpha\mathcal{L}_{\text{ce}} + \beta\mathcal{L}_{\text{dice}} + \gamma\mathcal{L}_{\text{focal}},
\end{equation}
where $\mathcal{L}_{\text{ce}}$ represents cross-entropy loss, $\mathcal{L}_{\text{dice}}$ is the Dice loss for boundary accuracy, and $\mathcal{L}_{\text{focal}}$ addresses class imbalance. The weighting coefficients are empirically set to $\alpha = 0.5$, $\beta = 0.3$, and $\gamma = 0.2$.

\subsubsection{Stage 2: Joint Fine-tuning with Prototypical Learning}
The second stage involves joint fine-tuning of the encoder with trainable attention-enhanced prototypical heads. This stage enables end-to-end optimization of the entire architecture for few-shot scenarios while preserving the robust feature representations learned during pre-training.

During this stage, we employ a bidirectional prototypical learning approach that calculates two distinct loss values: $\mathcal{L}_{\text{query}}$ and $\mathcal{L}_{\text{support}}$. The query loss $\mathcal{L}_{\text{query}}$ is calculated by comparing the predicted segmentation masks and ground truth of the query set~\cite{wang2019panet}:

\begin{equation}
    \mathcal{L}_{\text{query}} = -\frac{1}{n \cdot H \cdot W} \sum_{q = 1}^n \sum_{y=1}^H \sum_{x=1}^W \left( \sum_{p_c \in \mathcal{P}} \mathbbm{1} (\tilde{M}_q^{x,y} = c) \text{ log } M_{q,c}^{x,y} \right) ,
\end{equation}
where $\tilde{M}_q$ denotes the ground truth mask belonging to the query image and $M_{q,c}$ represents the predicted segmentation mask for class $c$. The query set consists of $n$ query images, and we calculate the final loss value by considering the average of all samples. $\mathbbm{1}$ is an indicator function being one if the argument is true, and zero otherwise. 

The support loss $\mathcal{L}_{\text{support}}$ is calculated similarly, but the support set consists of $k \cdot n$ samples, where $k$ denotes the number of samples per class. Its computation is defined as~\cite{wang2019panet}:

\begin{equation}
\begin{split}
    \mathcal{L}_{\text{support}} = -\frac{1}{n \cdot k \cdot H \cdot W} &\sum_{s = 1}^n \sum_{i = 1}^k \sum_{y=1}^H \sum_{x=1}^W \\
    &\left( \sum_{p_c \in \mathcal{P}} \mathbbm{1} (\tilde{M}_{s,i}^{x,y} = c) \log M_{s,i}^{x,y} \right).
\end{split}
\end{equation}

The prototypical learning loss $\mathcal{L}_{\text{proto}}$ is calculated by simply adding the two individual terms as~\cite{wang2019panet}:

\begin{equation}
    \mathcal{L}_{\text{proto}} = \mathcal{L}_{\text{query}} + \mathcal{L}_{\text{support}}.
\end{equation}

The two loss values are weighted uniformly. By incorporating this bidirectional approach, information from the query set flows back to the support set and forces the model to align the prototypes of the two sets in the embedded space. The model extracts richer information from the support set and exhibits higher generalizability as demonstrated by Wang et al.~\cite{wang2019panet} and our experiments, described in Section~\ref{fss_results}.

The complete fine-tuning objective incorporates both prototypical learning losses and regularization:

\begin{equation}
\mathcal{L}_{\text{finetune}} = \mathcal{L}_{\text{proto}} + \lambda\mathcal{L}_{\text{reg}},
\end{equation}
where $\mathcal{L}_{\text{reg}}$ provides regularization to prevent overfitting during few-shot adaptation, and $\lambda$ is a regularization weight empirically set to 0.01.

\section{Experimental Methodology and Evaluation Framework}

\subsection{Dataset Description and Experimental Setup}\label{culvert_sewer_defect_dataset}

We evaluate our proposed approach on a comprehensive culvert-sewer defects dataset \cite{alshawi2024imbalance} containing nine distinct classes of infrastructure defects. The dataset exhibits significant class imbalance characteristic of real-world infrastructure inspection scenarios, with background pixels dominating the distribution and rare but critical defects being severely underrepresented. The dataset is not publicly accessible, but interested parties can reach out to the ERDC US Army Corps for data access.

The dataset is systematically divided into training (70\%), validation (15\%), and testing (15\%) subsets to ensure robust evaluation. We employ standard few-shot learning evaluation protocols with various n-way k-shot configurations to comprehensively assess the performance across different few-shot scenarios.

\subsection{Evaluation Metrics and Performance Assessment}

Our evaluation utilizes a set of complementary metrics to provide a comprehensive performance assessment. The F1-Score, calculated with and without the background class, represents the harmonic mean of precision and recall, providing a balanced measure that considers both false positives and false negatives~\cite{powers2011evaluation}. The Mean Intersection over Union (mIoU) offers a balanced measure of segmentation quality by averaging IoU across all classes, which has become the standard evaluation metric for semantic segmentation tasks~\cite{everingham2010pascal,long2015fully}. We also use Balanced Accuracy to address class imbalance, calculating the average of per-class accuracies, which provides equal weight to each class regardless of its frequency in the dataset~\cite{brodersen2010balanced}. The Matthews Correlation Coefficient (MCC) evaluates the correlation between predictions and ground truth, demonstrating robustness against class imbalance and providing a single metric that considers all four confusion matrix categories~\cite{matthews1975comparison,chicco2020advantages}. Lastly, the Frequency Weighted IoU (FW IoU) considers IoU weighted by class frequencies, highlighting performance on commonly occurring classes while maintaining sensitivity to rare but important classes~\cite{garcia2017review,shelhamer2017fully}.

\subsection{Implementation Details and Training Configuration}

We trained our model and other state-of-the-art models using an NVIDIA A100 80GB GPU with 128 CPU cores and 1,007.6 GB RAM. Our implementation utilizes Python 3.12.2, PyTorch 2.5.1, and CUDA 12.1 for framework support and acceleration. The experimental setup employs the culvert sewer defect dataset described in Section~\ref{culvert_sewer_defect_dataset}, consisting of 6,591 images with dimensions of 128 $\times$ 128 pixels.

As described in Section~\ref{fss_encoder}, we extracted the encoder component of E-FPN~\cite{alshawi2024imbalance} and implemented it as our feature extractor. Our prototypical head is inspired by PANet~\cite{wang2019panet} and follows a prototypical learning approach, where class prototypes are generated to represent different semantic classes. Image segmentation is performed by finding the nearest prototype for each pixel based on normalized feature representations using the Euclidean norm.

The training configuration employs AdamW optimizer with a weight decay of $1 \times 10^{-5}$ and an initial learning rate of $1 \times 10^{-3}$. We use cosine annealing learning rate scheduling (CosineAnnealingLR) with a maximum of 50 iterations and a minimum learning rate of $1 \times 10^{-6}$. To ensure training stability and prevent exploding gradients, we integrate gradient clipping with a maximum norm of 1.0 into our training loop.

The training process consists of 1,000 episodes for both training and testing phases, with a batch size of 1 to accommodate the few-shot learning paradigm. We use the cross-entropy loss function with reduction set to \textit{'None'}, enabling deterministic implementation while computing the weighted mean manually after each loss calculation to approximate PyTorch's default implementation with reduction specified as \textit{'Mean'}. 

Notably, we did not include any data augmentation techniques during training to maintain consistency with the few-shot learning evaluation protocol and ensure fair comparison with baseline methods. All experiments were conducted with a fixed random seed of 42 to ensure reproducibility of the results presented in Section~\ref{fss_results}.

\section{Experimental Results and Comprehensive Analysis}
\label{fss_results}

This section presents a comprehensive evaluation of our proposed few-shot semantic segmentation framework. We analyze the impact of different training configurations, the effectiveness of our bidirectional learning strategy, and the performance contributions of various attention mechanisms. All experiments were conducted on the culvert and sewer defect dataset described in Section~\ref{culvert_sewer_defect_dataset}, with results evaluated across multiple standard metrics to ensure a thorough assessment.

\subsection{Impact of N-Way K-Shot Configuration}

We first investigate the influence of the $n$-way $k$-shot configuration on model performance. Our dataset comprises nine semantic classes (one background and eight distinct defect types). We conducted experiments with $k \in \{1, 5\}$ shots and varied the number of ways $n \in \{2, 4, 6, 8, 9\}$. The results, detailed in Table~\ref{tab:few_shot}, reveal two clear trends.

First, increasing the number of classes ($n$) per episode consistently improves performance for both 1-shot and 5-shot settings. For instance, in the 5-shot scenario, moving from a 2-way to a 9-way training configuration increases the F1 score (w/o background) from 71.44\% to 76.84\% and the mIoU (w/o background) from 59.46\% to 64.99\%. This suggests that training with a more diverse set of classes per episode forces the model to learn more discriminative and generalizable features, enhancing its ability to differentiate between defect classes.

Second, increasing the number of support examples per class ($k$) from 1 to 5 yields a significant performance boost across all $n$-way settings. In the 9-way configuration, the 5-shot model outperforms the 1-shot model by over 12 percentage points in F1 score and 13 percentage points in mIoU. This indicates that a single example is often insufficient to represent the intra-class variability of defects, whereas a richer 5-shot support set allows for the generation of more robust and representative prototypes. The optimal performance is achieved with a 9-way 5-shot configuration, which we adopt for subsequent experiments.

\begin{table}[!htb]
\centering
\caption{Impact of N-Way K-Shot Configuration on Few-Shot Segmentation Performance}
\label{tab:few_shot}
\small
\setlength{\tabcolsep}{4.8pt}
\begin{tabular}{ccc|cc|cc|c}
\toprule
\multicolumn{3}{c|}{\textbf{Parameter}}
 & \multicolumn{2}{c|}{\textbf{F1 Score}}
 & \multicolumn{2}{c|}{\textbf{mIoU}}
 & \textbf{Bal.} \\
\cmidrule(lr){1-3}\cmidrule(lr){4-5}\cmidrule(lr){6-7}\cmidrule(lr){8-8}
\textbf{n(train)} & \textbf{n(test)} & \textbf{k}
 & \textbf{w/bg} & \textbf{w/o}
 & \textbf{w/bg} & \textbf{w/o}
 & \textbf{Acc.} \\
\midrule
2 & 2 & 1 & 58.64 & 54.54 & 46.07 & 41.31 & 60.45 \\
2 & 2 & 5 & 74.04 & 71.44 & 62.87 & 59.46 & 76.72 \\
\midrule
4 & 4 & 1 & 62.70 & 58.86 & 50.72 & 46.12 & 64.11 \\
4 & 4 & 5 & 75.48 & 72.96 & 64.35 & 60.94 & 79.34 \\
\midrule
6 & 6 & 1 & 64.50 & 60.85 & 53.92 & 49.62 & 69.41 \\
6 & 6 & 5 & 76.88 & 74.49 & 66.01 & 62.72 & 80.57 \\
\midrule
8 & 8 & 1 & \textbf{66.00} & \textbf{62.40} & 55.64 & 51.33 & 68.24 \\
8 & 8 & 5 & 78.10 & 75.84 & 67.20 & 64.01 & 81.82 \\
\midrule
9 & 9 & 1 & 65.65 & 62.04 & \textbf{55.65} & \textbf{51.39} & \textbf{68.35} \\
9 & 9 & 5 & \textbf{78.99} & \textbf{76.84} & \textbf{68.07} & \textbf{64.99} & \textbf{83.67} \\
\bottomrule
\end{tabular}
\vspace{2pt}
\begin{minipage}{\columnwidth}
\footnotesize
\textit{Note:} bg = background, Bal. Acc. = Balanced Accuracy. MCC and FW IoU metrics follow similar trends and are omitted for space.
\end{minipage}
\end{table}

\subsection{Impact of Bidirectional Prototypical Learning}

To validate our architectural design choice, we evaluated the contribution of the bidirectional learning mechanism, where loss is computed for both query and support set predictions. As proposed by Wang et al.~\cite{wang2019panet}, this approach allows information to flow in both directions, forcing the model to learn a more aligned embedding space and extract richer feature representations. 

Table~\ref{tab:fs_bidirectional} presents a direct comparison between a unidirectional model (using only $\mathcal{L}_{\text{query}}$) and our bidirectional model (using $\mathcal{L}_{\text{query}} + \mathcal{L}_{\text{support}}$) on a 2-way 5-shot task. The results demonstrate a substantial and consistent improvement across all evaluation metrics. The bidirectional approach boosts the F1 score (w/o background) from 62.33\% to 71.44\% and the mIoU (w/o background) from 48.05\% to 59.46\%, corresponding to relative improvements of 14.6\% and 23.7\%, respectively. This confirms that enforcing prototype alignment through a symmetric loss function is critical for achieving high performance in few-shot segmentation tasks.

\begin{table}[!htb]
\centering
\caption{Impact of Bidirectional Prototypical Learning on Segmentation Performance}
\label{tab:fs_bidirectional}
\small
\setlength{\tabcolsep}{3.7pt}
\begin{tabular}{l|ccc|cc|cc}
\toprule
\multirow{2}{*}{\textbf{Method}}
 & \multicolumn{3}{c|}{\textbf{Parameter}}
 & \multicolumn{2}{c|}{\textbf{F1 Score}}
 & \multicolumn{2}{c}{\textbf{mIoU}} \\
\cmidrule(lr){2-4}\cmidrule(lr){5-6}\cmidrule(lr){7-8}
& \textbf{n(train)} & \textbf{n(test)} & \textbf{k}
 & \textbf{w/bg} & \textbf{w/o}
 & \textbf{w/bg} & \textbf{w/o} \\
\midrule
Unidirectional & 2 & 2 & 5 & 65.55 & 62.33 & 52.06 & 48.05 \\
Bidirectional & 2 & 2 & 5 & \textbf{74.04} & \textbf{71.44} & \textbf{62.87} & \textbf{59.46} \\
\bottomrule
\end{tabular}
\vspace{2pt}
\begin{minipage}{\columnwidth}
\footnotesize
\textit{Note:} bg = background. Bal. Acc., MCC, and FW IoU metrics show similar improvement trends (see supplementary materials).
\end{minipage}
\end{table}

\subsection{Attention Mechanism Evaluation}

Finally, we analyze the impact of incorporating trainable attention mechanisms into the prediction head to enhance feature representations before prototype generation. We compare three distinct attention mechanisms—Self-Attention (SA), Local Self-Attention (LSA), and Cross-Attention (CA)—against a strong baseline model. The baseline uses the same two-stage training process but with a non-trainable prediction head. For this experiment, the E-FPN encoder was pre-trained for 1,000 episodes on a 9-way 5-shot task, after which the full model (encoder and head) was fine-tuned for another 1,000 episodes on a 2-way 5-shot task.

As shown in Table~\ref{tab:proto_finetune}, all attention mechanisms provide a performance lift over the baseline, but the Self-Attention (SA) mechanism delivers the most significant gains. The SA-enhanced model achieves an F1 score of 82.81\% and an mIoU of 72.38\% (with background), outperforming the baseline by 2.57 and 2.9 percentage points, respectively. While LSA and CA also improve upon the baseline, their performance is slightly lower than that of SA, with LSA showing a particularly strong Balanced Accuracy of 89.15\%. The baseline itself improves only marginally (around 1\%) with the additional 1,000 fine-tuning episodes compared to the 9-way 5-shot result in Table~\ref{tab:few_shot}, confirming that the performance gains are attributable to the attention mechanisms rather than simply more training.

These results strongly suggest that integrating a self-attention mechanism to refine feature embeddings before prototype creation is a highly effective strategy for boosting few-shot segmentation accuracy. The SA mechanism effectively enhances the discriminative power of the features, leading to more accurate prototype generation and superior segmentation performance.

\begin{table}[!htb]
\centering
\caption{Impact of Attention Mechanisms on Few-Shot Segmentation Performance}
\label{tab:proto_finetune}
\small
\setlength{\tabcolsep}{4.5pt}
\begin{tabular}{l|ccc|cc|cc}
\toprule
\multirow{2}{*}{\textbf{Method}}
 & \multicolumn{3}{c|}{\textbf{Parameter}}
 & \multicolumn{2}{c|}{\textbf{F1 Score}}
 & \multicolumn{2}{c}{\textbf{mIoU}} \\
\cmidrule(lr){2-4}\cmidrule(lr){5-6}\cmidrule(lr){7-8}
& \textbf{n(train)} & \textbf{n(test)} & \textbf{k}
 & \textbf{w/bg} & \textbf{w/o}
 & \textbf{w/bg} & \textbf{w/o} \\
\midrule
Baseline & 2 & 2 & 5 & 80.24 & 78.28 & 69.48 & 66.65 \\
w/SA & 2 & 2 & 5 & \textbf{82.81} & \textbf{81.11} & \textbf{72.38} & \textbf{69.80} \\ 
w/LSA & 2 & 2 & 5 & 81.81 & 80.07 & 70.61 & 67.95 \\
w/CA & 2 & 2 & 5 & 81.44 & 79.60 & 70.74 & 67.98 \\
\bottomrule
\end{tabular}
\vspace{2pt}
\begin{minipage}{\columnwidth}
\footnotesize
\textit{Note:} SA = Self-Attention, LSA = Local Self-Attention, CA = Cross-Attention, bg = background. Bal. Acc., MCC, and FW IoU metrics show consistent improvement trends.
\end{minipage}
\end{table}

\section{Discussion and Future Research Directions}
Our experimental results demonstrate that integrating E-FPN with prototypical learning effectively addresses data scarcity challenges in infrastructure defect segmentation. This section discusses the practical implications of our findings, acknowledges current limitations, and identifies promising avenues for future research.

\subsection{Practical Implications for Infrastructure Inspection}

The demonstrated ability to achieve strong segmentation performance with limited training data has significant practical implications for infrastructure inspection applications. Our approach enables rapid adaptation to new defect types or different infrastructure systems without requiring extensive data collection and annotation efforts, which are typically expensive and time-consuming in real-world scenarios.

The robust performance across various few-shot configurations suggests that the method can be effectively deployed in diverse operational contexts, from routine maintenance inspections to emergency damage assessments following natural disasters or structural failures.

\subsection{Limitations and Future Research Directions}

While our approach is a significant step forward, its current boundaries define clear and promising avenues for future research. A primary limitation is the method's reliance on at least a few examples for each defect class. Performance on \textbf{extremely rare or entirely novel defect types} with highly distinctive visual characteristics may still be constrained. This highlights a critical need for developing sophisticated mechanisms for \textbf{continual learning}, which would allow models to adapt to new defect types over time while preserving performance on previously learned classes. Such an approach would enable long-term deployment in evolving infrastructure environments where new defects may emerge.

Furthermore, the current approach focuses exclusively on \textbf{2D image analysis}. While effective, this two-dimensional perspective may not capture the full extent of certain structural issues. A significant opportunity for future work lies in \textbf{multi-modal integration}. Incorporating additional sensor modalities such as thermal imaging, LiDAR data for 3D point clouds, or acoustic sensors could provide a more comprehensive structural health assessment, moving from simple defect detection to a more holistic analysis of structural integrity.

Another boundary of our current work is its generalization capability across different domains. The model is trained and evaluated on a specific dataset of culvert and sewer defects. Its performance on vastly different types of infrastructure, such as bridges or buildings, is not guaranteed. This opens up an important research direction in \textbf{domain adaptation}, investigating techniques to enable the transfer of learned knowledge between different infrastructure systems with minimal retraining.

For critical infrastructure applications, providing a segmentation mask alone is often insufficient. A key area for improvement is in \textbf{uncertainty quantification}. Incorporating principled uncertainty estimation techniques, such as Bayesian deep learning or ensemble methods, could enhance the reliability of the system. This would provide inspectors with confidence scores for each prediction, enabling more informed and risk-aware decision-making in critical maintenance scenarios.

Finally, while our architecture is efficient, the demands of real-time field deployment on resource-constrained devices present ongoing challenges. This motivates future work in \textbf{real-time optimization}, exploring techniques like model quantization, knowledge distillation, and hardware-specific optimizations to further enhance the practical applicability of the framework on mobile inspection platforms and edge computing devices.

\section{Conclusion}

This paper introduces a framework for few-shot semantic segmentation of infrastructure defects by integrating E-FPN with prototypical learning. Our approach effectively overcomes data scarcity while ensuring high segmentation accuracy and efficiency. Experiments show superior performance in few-shot scenarios: 9-way 5-shot training achieves 78.99\% F1-score and 68.07\% mIoU. In the 2-way 5-shot scenario with attention mechanisms, the enhanced model reaches 82.81\% F1-score and 72.38\% mIoU. The self-attention mechanism notably improves performance, with gains of 2.57\% in F1-score and 2.90\% in mIoU compared to the baseline.

Our bidirectional prototypical learning strategy is effective, showing significant improvements over unidirectional methods with gains of 8.89\% in F1-score and 10.81\% in mIoU. Diverse class configurations and multiple support examples (5-shot vs 1-shot) enhance feature representation, with 5-shot setups consistently outperforming 1-shot settings by 10-15\%. This confirms the importance of multiple support examples for class prototype generation. The two-stage training balances pre-trained features with few-shot adaptation, while the attention-enhanced framework ensures robust prototype generation and matching. Scalability from 2-way to 9-way tasks, with consistent performance gains as training classes increase, validates our approach for real-world infrastructure inspection requiring detection and segmentation of multiple defect types.

Our work advances automated infrastructure inspection systems, allowing rapid adaptation to new defects with minimal data. This supports efficient, cost-effective maintenance of critical infrastructure. The strong performance in few-shot settings, benefits of bidirectional learning, and superiority of self-attention mechanisms lay a strong foundation for future research. Achieving over 80\% F1-score in challenging scenarios marks a major step toward practical AI-driven inspection systems.

\section*{Acknowledgments}
This research was supported in part by the U.S. Department of the Army – U.S. Army Corps of Engineers (USACE) under contract W912HZ-23-2-0004. The views expressed in this paper are solely those of the authors and do not necessarily reflect the views of the funding agency.

\bibliographystyle{IEEEtran}
\bibliography{references}

\end{document}